\documentclass{article}

\usepackage[preprint]{neurips_2025}

\usepackage[utf8]{inputenc} 
\usepackage[T1]{fontenc}    
\usepackage{hyperref}       
\usepackage{url}            
\usepackage{booktabs}       
\usepackage{amsfonts}       
\usepackage{nicefrac}       
\usepackage{microtype}      
\usepackage{xcolor}         

\usepackage{graphicx}
\usepackage{subfigure}
\usepackage{hyperref}
\usepackage{amsmath}
\usepackage{amssymb}
\usepackage{mathtools}
\usepackage{amsthm}
\usepackage{wrapfig}
\usepackage{caption}
\usepackage{listings}
\usepackage{adjustbox}
\usepackage[capitalize,noabbrev]{cleveref}

\theoremstyle{plain}

\theoremstyle{definition}

\theoremstyle{remark}

\usepackage[textsize=tiny]{todonotes}
\usepackage{placeins}
\usepackage{siunitx}
\usepackage{enumitem}
\usepackage{float}

\title{ImageReFL: Balancing Quality and Diversity in Human-Aligned Diffusion Models}

\author{%
  Dmitrii Sorokin \\
  HSE University \\
  \texttt{storchrik@gmail.com} \\
  \And
  Maksim Nakhodnov \\
  AIRI \\
  \texttt{nakhodnov17@gmail.com} \\
  \And
  Andrey Kuznetsov \\
  AIRI, Sber, Innopolis \\
  \texttt{kuznetsov@airi.net} \\
  \And
  Aibek Alanov \\
  HSE University, AIRI \\
  \texttt{alanov.aibek@gmail.com} \\
}

\begin{document}

\maketitle

\begin{abstract}
Recent advances in diffusion models have led to impressive image generation capabilities, but aligning these models with human preferences remains challenging. Reward-based fine-tuning using models trained on human feedback improves alignment but often harms diversity, producing less varied outputs. In this work, we address this trade-off with two contributions. First, we introduce \textit{combined generation}, a novel sampling strategy that applies a reward-tuned diffusion model only in the later stages of the generation process, while preserving the base model for earlier steps. This approach mitigates early-stage overfitting and helps retain global structure and diversity. Second, we propose \textit{ImageReFL}, a fine-tuning method that improves image diversity with minimal loss in quality by training on real images and incorporating multiple regularizers, including diffusion and ReFL losses. Our approach outperforms conventional reward tuning methods on standard quality and diversity metrics. A user study further confirms that our method better balances human preference alignment and visual diversity. The source code can be found
at \href{https://github.com/ControlGenAI/ImageReFL}{github.com/ControlGenAI/ImageReFL}.
\end{abstract}

\section{Introduction}\label{sec:introduction}

Text-to-Image Diffusion Models \cite{dm,latent_diffusion_models} emerged as powerful tools for generating high-quality images; however, aligning them with human preferences remains a significant challenge. A similar issue in text generation was effectively addressed through Reinforcement Learning from Human Feedback (RLHF) \cite{rlhf,dpo,rl}, which led to substantial improvements in large language models. To reproduce such advances in image generation, several human preference datasets were introduced \cite{pick_a_pick, hps, image_reward}, enabling the training of models \cite{pick_a_pick, hps, mps, image_reward} that serve both as evaluation metrics and as reward functions for RL-based fine-tuning. Recent diffusion fine-tuning methods \cite{training_diffusion_rl, dpok, ddpo, apdm}, inspired by RLHF techniques such as DPO, optimized directly on preference data without relying on explicit reward models. Despite this conceptual simplicity, they showed only marginal improvements in standard human alignment metrics \cite{pick_a_pick, hps, mps, image_reward}, while requiring extensive computational resources. A more promising direction involves optimizing reward signals through direct backpropagation within the diffusion process, which significantly improves attractiveness metrics and offers better sample efficiency compared to policy gradient methods.



Following the idea of direct reward optimization, several works explored fine-tuning diffusion models by backpropagating through the reward function, in contrast to less efficient policy gradient methods. However, full backpropagation through the entire diffusion process is computationally expensive. To address this, ReFL algorithms~\cite{image_reward} restrict gradient flow to the final denoising step, detaching earlier steps from the computation graph. This idea has been adopted and extended in related approaches~\cite{dpok, alignprop, drtune}. While these methods significantly improve attractiveness metrics, they often lead to overfitting and reward hacking -- where the model exploits the reward function in unintended ways. As a result, diffusion models fine-tuned in this manner tend to produce visually similar outputs, suffering from diversity collapse and exhibiting novel visual artifacts not present in the base models.

Our work is motivated by the challenge of improving image quality without sacrificing diversity. While ReFL-like methods enhance visual appeal, they often lead to diversity collapse due to overfitting and reward hacking, whereas the original, non-fine-tuned models preserve diversity but fall short in matching human aesthetic and semantic preferences. To reconcile this trade-off, we leverage the observation that global image features are largely established in the early diffusion steps and remain stable in later stages~\cite{timesteps_1, timesteps_2}. Building on this insight, we propose combined generation, a strategy that uses the base model for the initial diffusion steps and switches to the reward-fine-tuned model only in the final steps. This approach improves image quality while retaining diversity, and enables explicit control over the quality–diversity balance by varying the handover point. However, as more steps are allocated to the base model, quality may deteriorate. To overcome this, we introduce ImageReFL -- a training method that fine-tunes the model to refine images using only the final diffusion steps. By adding standard diffusion loss as a regularizer and applying the ReFL loss across multiple timesteps, our approach mitigates overfitting and enhances robustness, allowing for high-quality generation with minimal compromise on diversity.

We evaluate our method using both image quality and diversity metrics. For quality, we report HPSv2.1, MPS, PickScore, and ImageReward~\cite{hps, mps, pick_a_pick, image_reward}. To assess diversity, we use FID~\cite{FID}, CLIPScore~\cite{clip} (reflecting text-image alignment), Covariance Distance, and Log Covariance Distance~\cite{rew_tradeoff}. Additionally, we propose DinoDiversity, a variant of CLIPDiversity~\cite{style}, which computes the average cosine distance between DINO~\cite{dino_1, dino_2} embeddings of samples generated from the same prompt. A user study comparing ReFL and ImageReFL confirms that our method improves diversity, textual alignment, and visual appeal.

Our main contributions are:

\begin{itemize}
\item We propose a combined generation strategy that uses the base model for early diffusion steps and the reward fine-tuned model in later steps, balancing quality and diversity.

\item We introduce ImageReFL, a fine-tuning method that refines real images using the final diffusion steps, combining reward-based and standard diffusion losses to improve quality while preserving diversity.

\item We provide comprehensive empirical evidence that ImageReFL outperforms prior methods across multiple quality and diversity metrics, as well as in human evaluations.
\end{itemize}

\section{Related Work}\label{sec:related_work}

\textbf{Text-to-image model alignment.} Recent work explores various approaches for aligning diffusion model outputs with human preferences. Early methods applied classical reinforcement learning to optimize reward functions~\cite{training_diffusion_rl, dpok}, followed by techniques that fine-tuned models directly on preference datasets without explicit reward modeling~\cite{ddpo, apdm}. Despite their conceptual appeal, these approaches showed only modest gains on standard alignment metrics~\cite{pick_a_pick, hps, image_reward, mps}. In parallel, other studies proposed optimizing reward functions via direct gradient backpropagation. A notable example is ImageReward~\cite{image_reward}, which introduced ReFL -- a fine-tuning method that restricts gradient flow to only the final diffusion step to reduce computational cost. This principle of truncated backpropagation was further explored in subsequent methods~\cite{alignprop, draftk, drtune}, showing improved reward alignment compared to RL-based approaches. However, these methods often overfit to the reward signal, leading to reward hacking, reduced diversity, and recurring visual artifacts. Our method addresses these limitations by preserving diversity while maintaining strong alignment with human preferences.

\textbf{Image quality evaluation.} Image quality assessment remains a complex task, involving both obvious flaws -- like anatomical distortions -- and subtler perceptual attributes that are difficult to formalize. Recent studies show that deep learning models trained on human preference data can effectively approximate these judgments~\cite{mps, hps, pick_a_pick, image_reward}, and such reward models are now commonly used for evaluation. However, optimizing for these metrics often leads models to overfit, resulting in artifacts such as unnatural grain, repetitive color schemes, or stylized lighting. These issues highlight the risks of directly optimizing reward functions. Our method mitigates these effects by incorporating the original, non–fine-tuned model during early diffusion steps, which helps preserve diversity and avoid undesirable visual artifacts.

\section{Background}

\subsection{Diffusion models}
Diffusion models generate images by reversing a gradual noising process. In the forward process, the clean latent $x_0$ is corrupted over $T$ steps by adding Gaussian noise:\\ $q(x_t | x_{t-1}) = \mathcal{N}(x_{t}; \sqrt{\alpha_t} x_{t-1}, (1 - \alpha_t)\mathbf{I})$, where $\alpha_t \in (0, 1]$ defines the noise schedule. This process can be rewritten as $x_t = \sqrt{\overline\alpha_t}x_0 + \sqrt{1 - \overline\alpha_t}\,\epsilon$, where $\epsilon \sim \mathcal{N}(0, \mathbf{I})$ and $\overline\alpha_t = \prod_{i=1}^{t} \alpha_i$.

The reverse process learns to predict the noise via $\epsilon_\theta(x_t, t, c)$, and generates previous sample:\\ $x_{t-1} = a_t x_t + b_t \epsilon_\theta(x_t, t, c) + c_t \hat{\epsilon}$, where $a_t$, $b_t$, and $c_t$ are sampling coefficients, $c$ is a text caption and $\hat{\epsilon} \sim \mathcal{N}(0, \mathbf{I})$ adds stochasticity.

The model is trained by minimizing the noise prediction loss:

\begin{equation}
    \mathcal{L}_{\text{diff}} = \mathbb{E}_{x_0, \epsilon, t} \left[ \left\| \epsilon - \epsilon_\theta(x_t, t, c) \right\|^2 \right].
\end{equation}

At inference time, a sample is generated by denoising a noise vector $x_T \sim \mathcal{N}(0, \mathbf{I})$ through $T$ steps.

\subsection{Reward fine-tuning}
Recent studies have demonstrated that diffusion models can be fine-tuned to directly optimize a reward model via standard gradient descent~\cite{image_reward, draftk, alignprop, drtune}. Although the specific implementations of these algorithms vary, they all follow a common principle: during training, a full denoising trajectory is generated, after which the resulting image $x_0 = \mathcal{F}_\theta(x_T, T, c)$ and the original prompt $c$ are passed to a reward function $R$. The output of the reward model is then used as a (negated) loss signal:

\begin{equation}
    \mathcal{L}_{\text{reward}} = -R(\mathcal{F}_\theta(x_T, T, c), c),  
\end{equation}

where $R(\cdot, \cdot)$ denotes the scalar reward predicted by the reward model, $ x_T \sim \mathcal{N}(0, 1)$.

However, performing gradient descent through the entire diffusion trajectory is computationally expensive due to the large number of steps and the increasing memory requirements. Consequently, most implementations freeze the model weights during the initial stages of the diffusion process, preventing gradients from propagating through the early denoising steps. ReFL~\cite{image_reward} performs a random number of diffusion steps during training, with $t_f \in [T_{\text{min}}, T_{\text{max}}]$, and backpropagates gradients only through the final step. It was empirically shown that the range $[T_{\text{min}}, T_{\text{max}}]$ should lie near the end of the reverse diffusion process -- typically $[0, 10]$ for a 40-step schedule.
Other training modifications were proposed \cite{draftk, alignprop, drtune}, but ReFL remains one of the most effective approaches, demonstrating significant improvements in image attractiveness.


\section{Method}\label{sec:methodology}

\subsection{Motivation}

\begin{figure}[tb]
    \includegraphics[width=\textwidth]{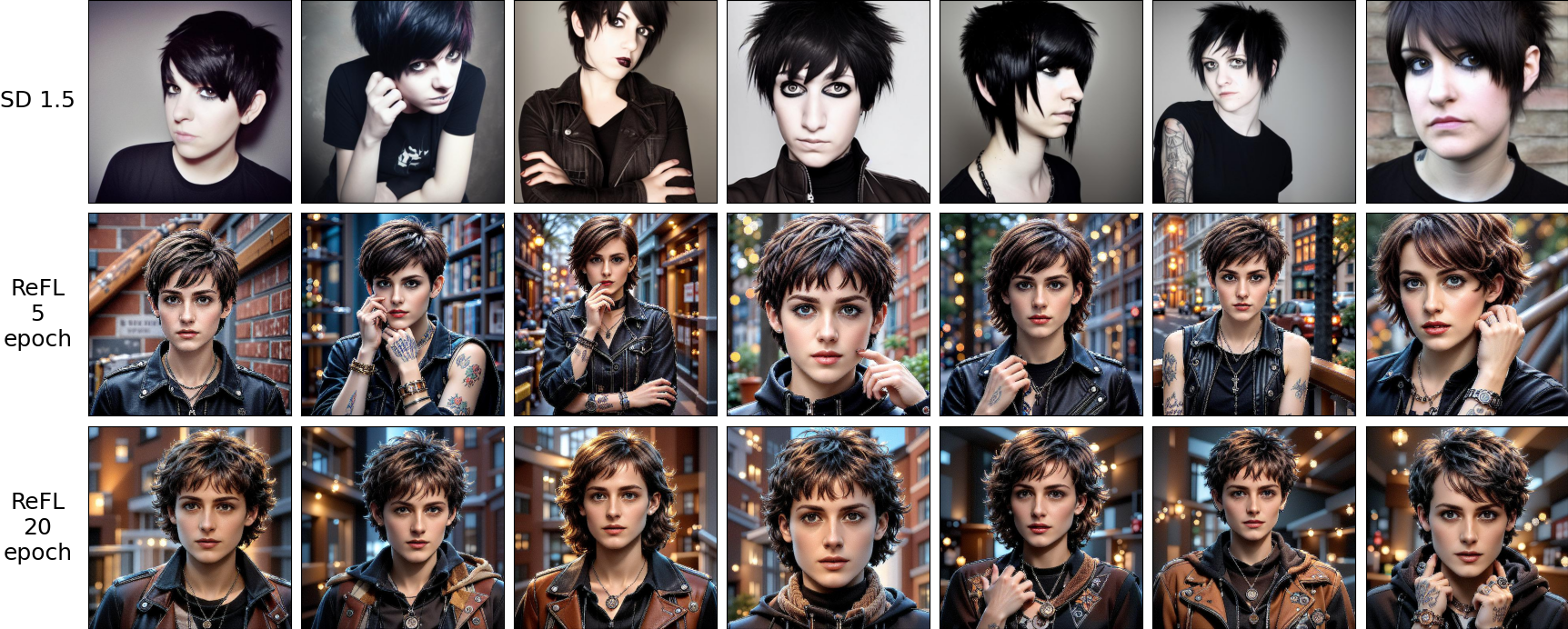}
    \caption{Diversity degradation during ReFL training on HPSv2.1 using SD1.5. The first row shows outputs from the original model, which produces diverse but less appealing images. As training progresses (second and third rows), visual appeal improves, but diversity collapses, with increasingly similar outputs across prompts.
    }
    \label{fig:fig_1}
\end{figure}

\begin{figure}[tb]
    \includegraphics[width=\textwidth]{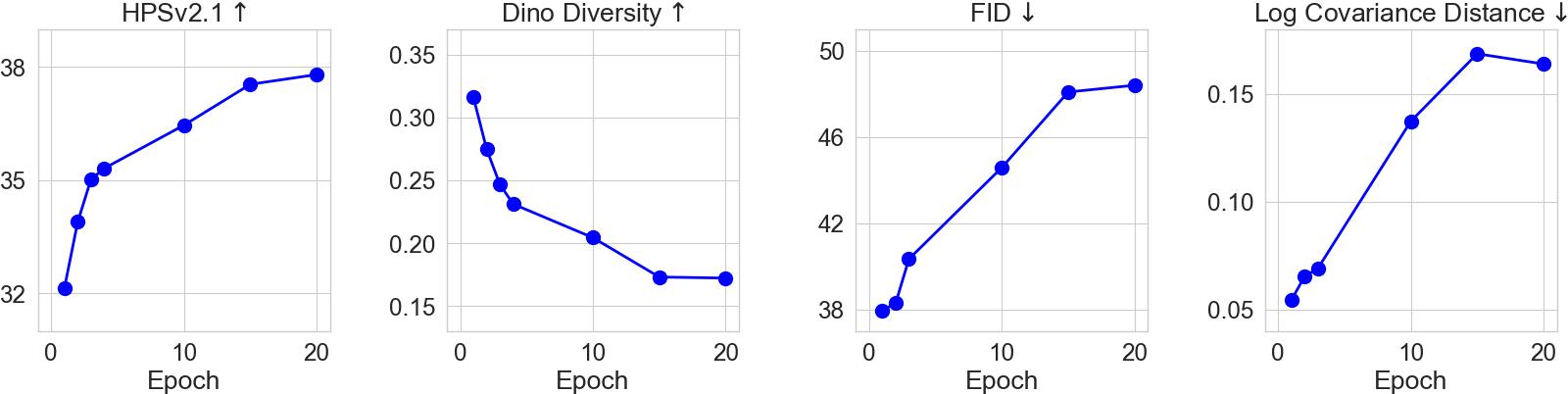}
    \caption{
        Image quality and diversity metrics during ReFL training on SD1.5 with HPSv2.1 as the target reward. HPSv2.1 indicates improvements in visual quality, while DinoDiversity, FID, and LogCovDistance capture the decline in diversity over training iterations.
    }
    \vspace{-1.3em}
    \label{fig:fig_2}
\end{figure}

ReFL demonstrates a strong ability to improve image quality. However, reward fine-tuned models often learn specific generative patterns that do not necessarily enhance the perceptual quality of the output but still increase the corresponding metric scores. Figure~\ref{fig:fig_1} illustrates the degradation in sample diversity throughout the training process. As training progresses, the ReFL fine-tuned model begins to generate visually similar faces and introduces consistent background artifacts, such as localized lighting spots. This behavior is indicative of the model overfitting to the reward function. In addition, the diversity metrics presented in Figure~\ref{fig:fig_2} show a clear decline in diversity (Dino Diversity, FID and Log Covariance Distance metrics) as the target quality metric (HPSv2.1) improves.

This behavior reflected the well-known issue of reward hacking, commonly observed in reinforcement learning. In the context of ReFL-based training, this issue became particularly severe, as it not only affected reward alignment but also significantly reduced the diversity of generated samples.

Since the entire generation trajectory is produced using the fine-tuned model, these patterns may influence both global image features -- such as composition and human pose -- and more local characteristics. It is known that global image attributes are typically formed during the early stages of the diffusion process and remain relatively stable in later steps\cite{timesteps_1, timesteps_2}. Therefore, if the fine-tuned model is applied only starting from a certain step of the diffusion trajectory, it becomes possible to preserve global image diversity. This insight led to our first methodological contribution -- combined generation.

\subsection{Combined generation}


To preserve the diversity of global image features -- such as composition, background, and human pose -- which are primarily formed during the early stages of the diffusion process, we propose a combined generation method. In this approach, we use the predictions of the original model $\epsilon_{\theta_0}(x_t, t, c)$ for the initial diffusion steps, and switch to the fine-tuned model $\epsilon_{\theta}(x_t, t, c)$ in the later steps:

\begin{equation}
    \epsilon_{combined}(x_t, t, c) =
    \begin{cases} 
        \epsilon_{\theta_0}(x_t, t,  c), & \text{if } t > T', \\
        \epsilon_{\theta}(x_t, t, c), & \text{if } t \leq T',
    \end{cases}\quad
    t = T,\dots,0
\end{equation}
$T' \in [0, T]$ is a hyperparameter of generation method, $\theta$ - weights of fine-tuned model, $\theta_0$ is the weights of original model.

\begin{figure}[tb]
    \vspace{-1em}
    \includegraphics[width=\textwidth]{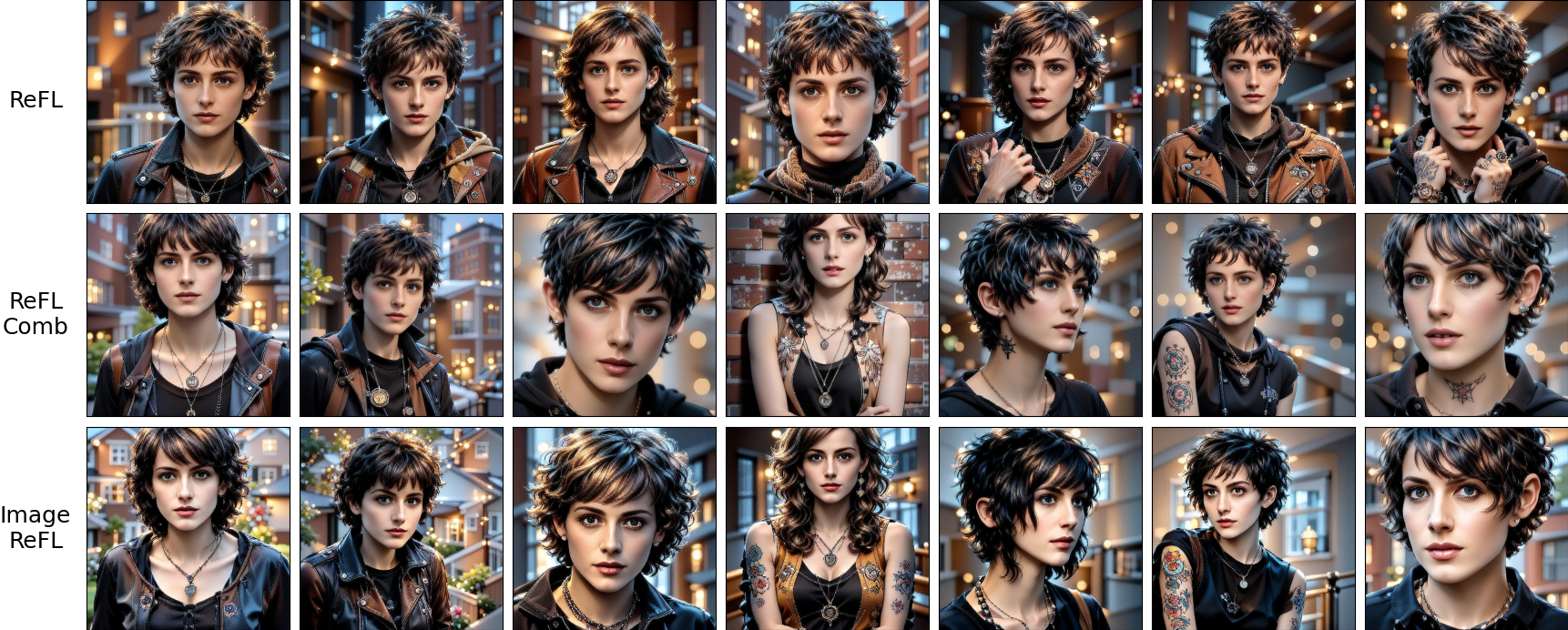}  
    \caption{Comparison of generations by ReFL, ReFL Combined, and ImageReFL. ReFL achieves high visual quality but suffers from low diversity. ReFL Combined improves diversity by using the base model in early steps, while ImageReFL further enhances the quality–diversity trade-off.
    }
    \vspace{-1.2em}
    \label{fig:fig_3}
\end{figure}
\begin{figure}[tb]
    \vspace{-1em}
    \includegraphics[width=\textwidth]{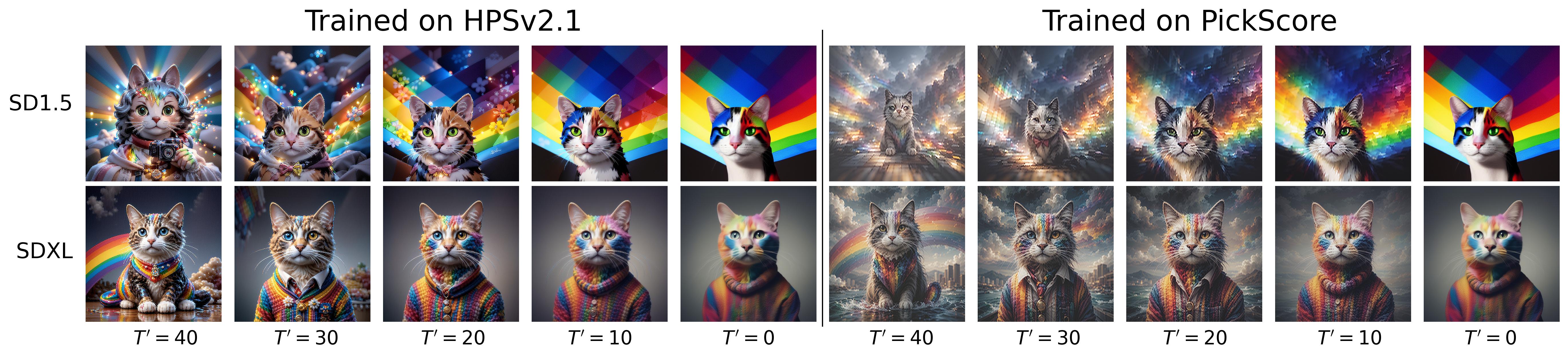}
    \caption{
    Demonstration of the combined generation method applied to different base models (Stable Diffusion 1.5 and SDXL) and target reward functions (HPSv2.1 and PickScore). 
    }
    \label{fig:fig_4}
\end{figure}

Figure~\ref{fig:fig_3} illustrates the effect of combined generation compared to standard ReFL. In this example, we apply the original (non–fine-tuned) model for the first 10 steps of the 40-step diffusion process and switch to the reward fine-tuned model for the final 30 steps. This setup results in noticeably more diverse outputs without sacrificing visual quality. Figure~\ref{fig:fig_4} further demonstrates how image generation depends on the choice of $T'$.

Figure~\ref{fig:fig_5} explores how varying the switch point in combined generation affects the quality–diversity trade-off. Specifically, we experiment with different values of $T' \in \{37, 35, 33, 30, 25, 20, 15, 8, 5\}$, where $T'$ denotes the step at which we switch from the original model to the reward fine-tuned model. Since standard ReFL requires a large number of denoising steps to significantly improve image appeal, we sample more configurations in this range. The results show that increasing the number of fine-tuned steps generally enhances image quality but reduces diversity, enabling explicit control over the trade-off. Compared to standard ReFL, which applies the fine-tuned model across all 40 steps, combined generation achieves a better balance between perceptual quality and sample diversity.


\subsection{ImageReFL}

As shown by our experiments, when applying the combined generation method to a model trained using standard ReFL, it was possible to achieve high image quality only when a sufficiently large number of diffusion steps -- typically at least half -- were performed using the fine-tuned model. As a result, deeper image features were also generated with the fine-tuned model, which reduced diversity. 

This behavior is likely due to the use of the reward fine-tuned model for the entire denoising trajectory during training. As a result, the model may apply reward-optimizing modifications already in the early diffusion steps, reducing the need for significant changes in the final steps. Consequently, it implicitly affects earlier stages of the generation process despite the restriction of gradient flow, effectively optimizing the reward signal beyond the final denoising steps.
Therefore, we propose an algorithm that explicitly prevents the fine-tuned model from influencing the early stages of image generation. Instead of generating the full denoising trajectory with the fine-tuned model, we train it to modify real images to increase their reward score:\\
In contrast to the standard ReFL algorithm, which performs the entire diffusion process starting from step $T$, our method begins from an intermediate latent representation $x_{t'}$, obtained by adding noise to a real image $x_0^{\text{real}}$ up to step $t'$:
$x_{t'} = \sqrt{\overline{\alpha}_{t'}} x_{0}^{\text{real}} + \epsilon \sqrt{1 - \overline{\alpha}_{t'}}$,
where $\epsilon \sim \mathcal{N}(0, I)$ is Gaussian noise. We then continue the diffusion process from step $t'$ to step $t_f$, after which the final latent representation $x_0$ is sampled. The entire sampling process with the trainable model starting from step $t'$ is denoted as
$x_0 = \mathcal{F}_\theta(x_{t'}, t', c)$.
Gradients are backpropagated only through the final diffusion step from $t_f$ to $t_0$, similarly to the original ReFL method. The reward loss is then defined as:

\begin{equation}
    \mathcal{L}_{\text{reward}}^{'} = -R(\mathcal{F}_\theta(x_{t'}, t', c), c)
\end{equation}

However, training solely with the reward-based loss $\mathcal{L}_{\text{reward}}'$ leads to severe overfitting, which negatively impacts image quality. To mitigate this issue, we introduce two forms of regularization.

First, we incorporate the standard diffusion loss to encourage the model to hold generative stability:

\begin{equation}
    \mathcal{L}_{\text{diffusion}} = \left\|\epsilon - \epsilon_\theta(x_{t'}, t', c)\right\|_2^2,
\end{equation}

Second, every 4 training steps, we apply the classical ReFL loss by generating full denoising trajectories. This regularization encourages the model to remain compatible with earlier diffusion steps and helps prevent overfitting to the reward signal by avoiding excessive specialization to only the final stages of generation.

\begin{figure}[tb]
    \vspace{-0em}
    \centering
    \includegraphics[width=\textwidth]{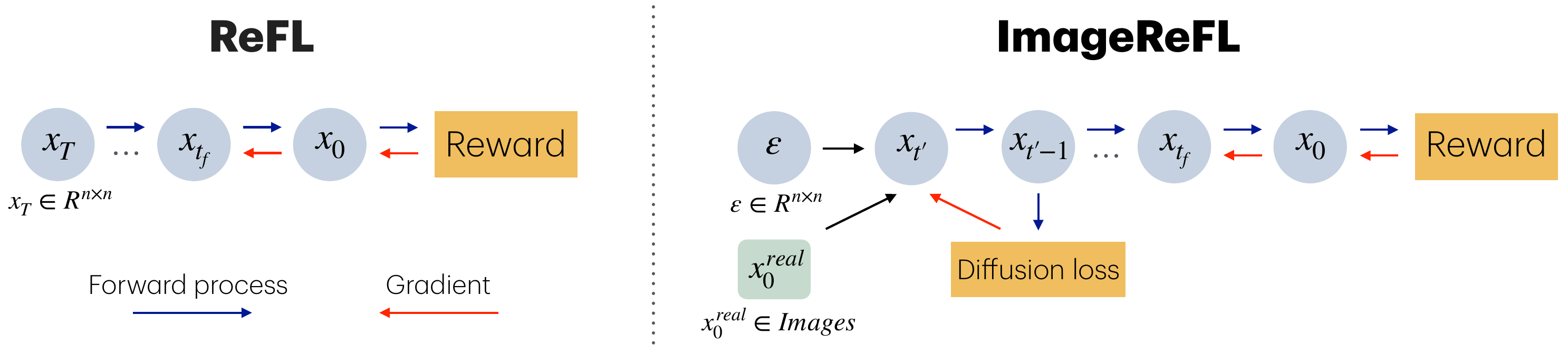}
    \caption{Comparison of ImageReFL with the standard approach. 
Unlike classical ReFL, ImageReFL incorporates real images to train the model to make significant changes during final diffusion steps.}
    \label{fig:fig_6}
\end{figure}

Figure~\ref{fig:fig_6} illustrates the key differences in the training process between the original ReFL algorithm and our modified approach, ImageReFL.

\begin{figure}[tb]
    \includegraphics[width=\textwidth]{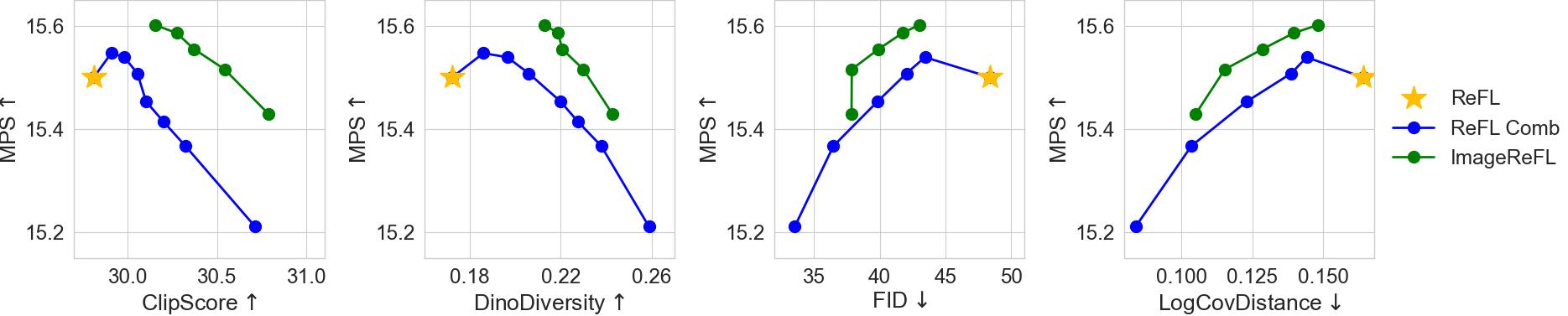}
    \caption{
        Image quality and diversity trade-off for SD1.5 trained on the HPSv2.1 score using combined generation. MPS reflects image quality, while DinoDiversity, FID, and LogCovDistance measure diversity.
    }
    \vspace{-1em}
    \label{fig:fig_5}
\end{figure}

\begin{figure}[tb]
    \centering
    \includegraphics[width=\textwidth]{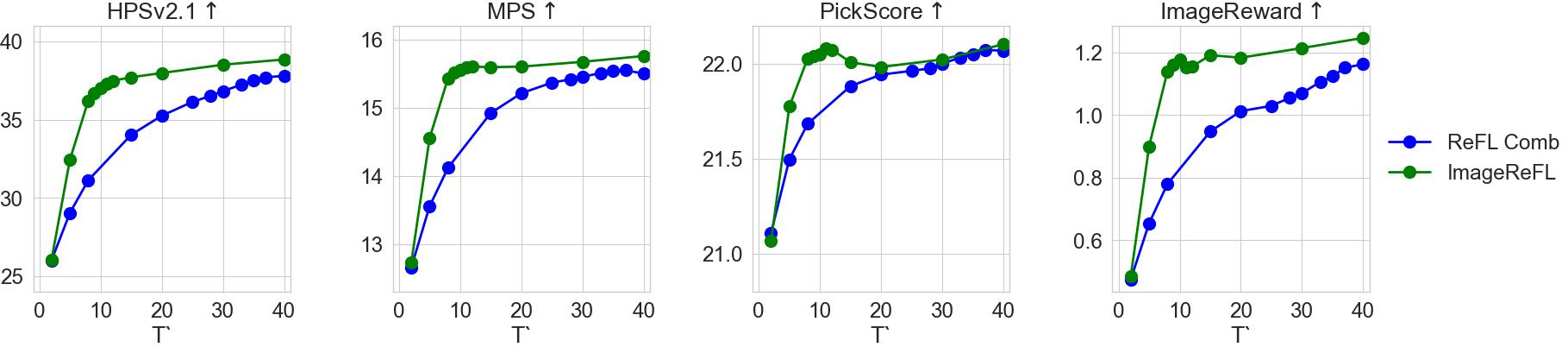}
    \caption{Comparison between ReFL and ImageReFL with combined generation. The X-axis indicates the diffusion step after which the fine-tuned model is applied. The Y-axis shows various image quality metrics.
    }
    \label{fig:steps_diversity}
\end{figure}

Figure~\ref{fig:steps_diversity} demonstrates the ability to significantly improve image appeal at several diffusion steps. While classical ReFL with combined generation requires at least 30 out of 40 steps using a fine-tuned model to produce high-quality images, ImageReFL achieves comparable or superior quality with only 10 steps.
In terms of the quality/diversity trade-off (Figure~\ref{fig:fig_5}), ImageReFL also outperforms the original ReFL with combined generation.

\section{Experiments}\label{sec:experimental_setup}

\subsection{Evaluation Metrics}

\textbf{Image quality.} To evaluate visual quality, we rely on four established reward-based metrics: Human Preference Score (HPS)~\cite{hps}, Multi-Preference Score (MPS)~\cite{mps}, PickScore~\cite{pick_a_pick}, and ImageReward~\cite{image_reward}. HPS and MPS demonstrate strong alignment with human preferences. PickScore and ImageReward are trained on different datasets and show lower correlation with HPS and MPS, which allows them to provide additional information -- especially useful when a model overfits to a reward.

\textbf{Diversity.} Measuring diversity remains a challenge, so we use a combination of metrics capturing different aspects. Frechet Inception Distance (FID)~\cite{FID} measures global similarity between real and generated image distributions but is not sensitive to prompt-level variation. Spectral Covariance Distance and Log Spectral Covariance Distance~\cite{rew_tradeoff} compare the spread of features via covariance matrices, providing insight into overall feature variability. DinoDiversity captures intra-prompt diversity by computing the average pairwise cosine distance between DINO~\cite{dino_1, dino_2} embeddings for images generated from the same prompt -- higher values indicate greater variation. Lastly, although CLIPScore~\cite{clip} is primarily a text-image alignment metric, it serves as a useful proxy: a lack of diversity often correlates with reduced alignment across generations.

\subsection{General Implementation Details}

As base models for our experiments, we selected Stable Diffusion 1.5 (SD1.5)~\cite{Rombach_2022_CVPR} for its relatively small size and computational efficiency, and Stable Diffusion XL (SDXL) for its superior image quality. For the reward signal during training, we employed two quality assessment metrics: HPSv2.1 and PickScore, each used as the optimization target in separate training runs.

\subsection{Training setup}

For all experiments, we use the same hyperparameters. We used the train split of the COCO dataset~\cite{coco} as the primary data source, both for the standard ReFL (where only the prompts from the dataset were used) and for ImageReFL. We employ the AdamW optimizer with a learning rate of $3 \times 10^{-4}$. 
The effective batch size is 32 for SD1.5 and 16 for SDXL training. 
We fine-tune only the U-Net weights in both models. 
All reward models are rescaled to the $[0, 1]$ range and further multiplied by $10^{-3}$. 
The diffusion loss is scaled by $10^{-5}$ for all training setups. 
For ImageReFL, we alternate training by applying one ReFL step after every three ImageReFL steps.

For the ReFL algorithm, we train for 20 full epochs, and for ImageReFL, we train for 40 epochs. 
Training takes approximately 9 hours for SD1.5 and 40 hours for SDXL (both for ReFL and ImageReFL) on a single NVIDIA A100 80GB GPU.

\begin{figure}[tb]
    \includegraphics[width=\textwidth]{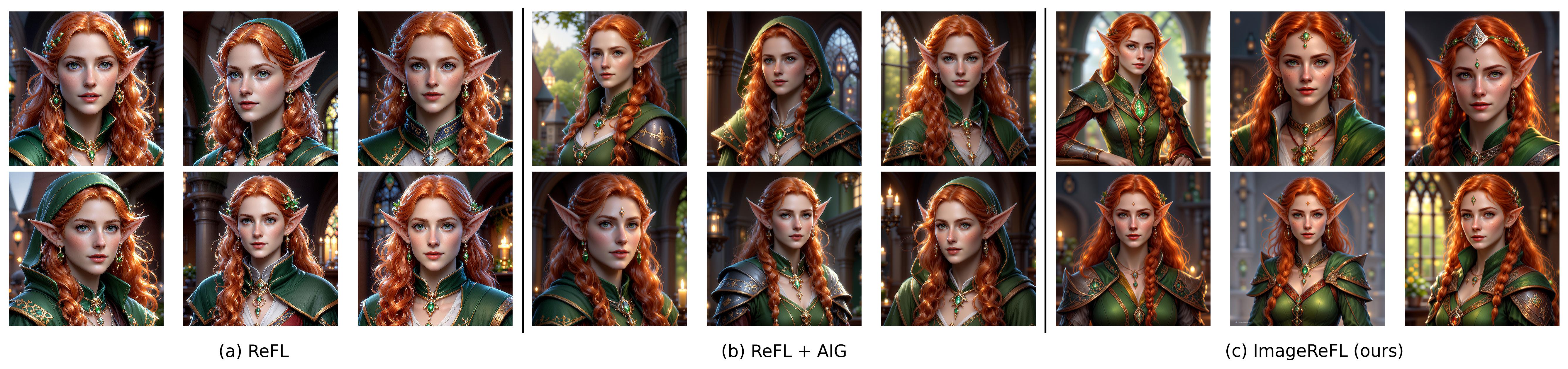}
    \caption{Comparison of ImageReFL with ReFL and ReFL AIG baselines using SDXL and HPSv2.1 as the target reward. All samples are generated with the same prompt. ImageReFL produces more appealing and diverse outputs, while baselines exhibit artifacts or mode collapse.}
    \label{fig:fig_8}
\end{figure}
\subsection{Combined inference setup}
To compute image quality metrics, we generate images using 400 prompts from the test split of the HPDv2 dataset. For each prompt, we generate 5 images and average the resulting scores. The same setup is used to compute DinoDiversity, which evaluates intra-prompt variation.
For dataset-level diversity metrics -- FID, CovDistance, and LogCovDistance -- we follow standard practice and use the test and validation splits of the MS COCO dataset (10,000 samples) as the reference distribution.

For both model configurations, we apply 40 denoising steps during inference. 
Among these, the first $n$ steps are performed using the original model, and the remaining $40 - n$ steps with the fine-tuned model. 
The original model is run with classifier-free guidance, as its performance degrades significantly without it. We use the default guidance scales: 7.5 for SD1.5 and 5.0 for SDXL. 
In contrast, the fine-tuned models achieve the best results without classifier-free guidance.

For SD1.5, we use 10 steps with the original model and 30 steps with the fine-tuned model for ReFL, while for ImageReFL, we use 30 original and 10 fine-tuned steps. For SDXL, we use 20 original and 20 fine-tuned steps for ReFL, and 35 original and 5 fine-tuned steps for ImageReFL.

\subsection{Metrics results}




We evaluate our approach across four experimental setups: SD1.5 trained with HPSv2.1, SD1.5 trained with PickScore, SDXL trained with HPSv2.1, and SDXL trained with PickScore. The results for models trained on HPSv2.1 are summarized in Table~\ref{tab:metrics_table} and graphs~\ref{fig:fig_9} \ref{fig:fig_10}, results for PickScore can be found in the appendix \ref{sec:appendix_pick_score_results}. 
For each configuration, we compare against two baselines: the original ReFL method and ReFL with Artificially-Inspired Guidance (ReFL AIG). These are evaluated alongside our proposed ImageReFL and ReFL with combined generation.

\begin{table}[t]
    \centering
    \caption{Unified image generation comparison across SD1.5 and SDXL trained with HPSv2.1.}
    \label{tab:metrics_table}
    \resizebox{\linewidth}{!}{
    \begin{tabular}{@{}llccccccccc@{}}
    \toprule
    \multicolumn{9}{c}{\textbf{Quality}} & {\textbf{Diversity}}\\
    \cmidrule(lr){3-7} \cmidrule(lr){8-11}
    Setting & Method &
    \text{HPSv2.1} $\uparrow$ &
    MPS $\uparrow$ &
    PickScore $\uparrow$ &
    IR $\uparrow$ &
    ClipScore $\uparrow$ &
    DinoDiv $\uparrow$ &
    FID $\downarrow$ &
    CovDist $\downarrow$ &
    LogCovDist $\downarrow$ \\
    
    & & $\pm$0.07 & $\pm$0.10 & $\pm$0.03 & $\pm$0.018 & $\pm$0.09 
    & $\pm$0.003 & $\pm$0.5 & $\pm$0.0002 & $\pm$0.001 \\
    \midrule

    \shortstack[l]{SD1.5} 
    & ReFL & \textbf{37.80} & \underline{15.50} & \textbf{22.07} & 1.163 & 29.81 & 0.172 & 48.4 & 0.0093 & 0.164 \\
    & ReFL AIG & 36.07 & 14.94 & 21.89 & 1.024 & 29.56 & \bf{0.231} & 42.5 & 0.0070 & \bf{0.102} \\
    \cmidrule(lr){2-11}
    & ReFL Comb & 36.83 & 15.45 & 22.00 & \underline{1.070} & \underline{30.10} & \underline{0.220} & \textbf{39.83} & \underline{0.0066} & \underline{0.123} \\
    & ImageReFL & \underline{37.00} & \textbf{15.55} & \underline{22.05} & \textbf{1.178} & \textbf{30.37} & \underline{0.220} & \underline{39.91} & \textbf{0.0059} & 0.129 \\

    \midrule
    \shortstack[l]{SDXL} 
    & ReFL & \textbf{39.31} & \textbf{16.36} & 22.96 & \underline{1.475} & 31.96 & 0.1532 & 45.3 & 0.0073 & 0.147 \\
    & ReFL AIG & \underline{37.48} & 15.71 & 23.00 & 1.295 & 31.86 & \bf{0.233} & 42.16 & 0.0075 & 0.093 \\
    \cmidrule(lr){2-11}
    & ReFL Combined & 37.02 & \underline{16.32} & \underline{23.25} & 1.445 & \textbf{33.69} & \underline{0.2317} & \textbf{29.0} & \textbf{0.0052} & \textbf{0.059} \\
    & ImageReFL & 37.00 & \underline{16.32} & \textbf{23.54} & \textbf{1.519} & \underline{33.59} & 0.2116 & \underline{31.7} & \underline{0.0054} & \underline{0.061} \\

    \bottomrule
    \end{tabular}
}
\vspace{-1em}
\end{table}
\hfill

In all setups, both ImageReFL and ReFL with combined generation consistently and significantly outperform the original ReFL across all diversity-related metrics, including DinoDiversity, FID, CovDist, and LogCovDist. These results demonstrate the effectiveness of our methods in generating more diverse visual outputs. The advantage of diversity is also demonstrated in the example of image generation (Figure~\ref{fig:fig_8}). Compared to ReFL AIG, our methods achieve significantly better alignment with textual descriptions, as indicated by higher CLIPScore values. Additionally, at a fixed level of image diversity, our methods produce substantially higher image quality across all metrics -- except for the primary training objective metric (HPSv2.1 or PickScore). 
\begin{wraptable}{r}{0.52\textwidth}
\centering
    \vspace{-0.7em}
    \caption{Win rate statistics comparing ReFL with ImageReFL across different setups.}
    \label{tab:user_study}
    \resizebox{\linewidth}{!}{
        \begin{tabular}{lcccc}
            \toprule
            & \multicolumn{2}{c}{\textbf{ReFL SD1.5}} & \multicolumn{2}{c}{\textbf{ReFL SDXL}} \\
            \cmidrule(lr){2-3} \cmidrule(lr){4-5}
            & HPS & PickScore & HPS  & PickScore \\
            \midrule
            Diversity    & 74.0 & 74.13 & 68.5 & 77.25  \\
            Accordance    & 51.8 & 56.45 & 70.7 & 60.8 \\
            Attractiveness   & 45.9 & 73.33 & 72.3 & 68.8 \\
            \bottomrule
    \end{tabular}
    }
    \vspace{-2.1em}
\end{wraptable} 
However, this discrepancy likely reflects overfitting to the reward function rather than a genuine reduction in perceptual quality.
Regarding quality, ImageReFL consistently matches or exceeds all baselines across most evaluation metrics (HPSv2.1, MPS, ImageReward, PickScore). The only exceptions are the training reward metrics, where slight degradation is expected due to baseline method overfitting.



\begin{figure}[tb]
    \includegraphics[width=\textwidth]{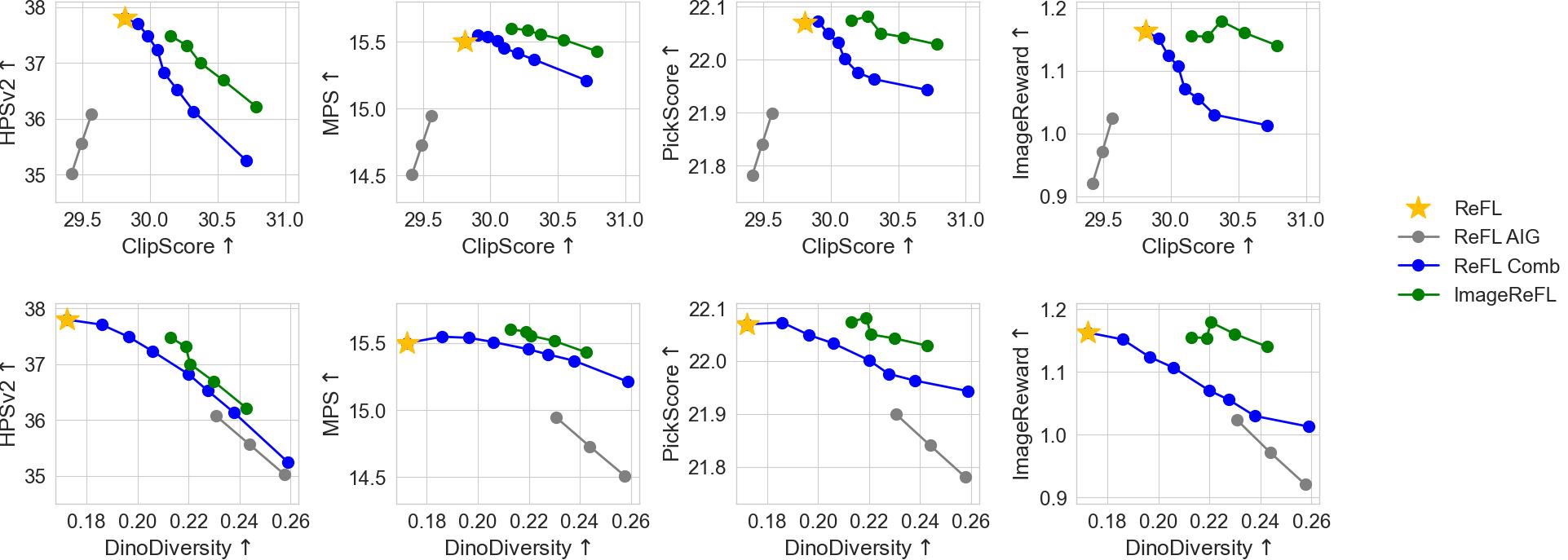}
    \caption{Quality/diversity trade-off for SD1.5 trained on HPSv2.1. ImageReFL outperforms both baselines and achieves a more favorable balance between image quality and diversity.}
    \label{fig:fig_9}
\end{figure}

\begin{figure}[tb]
    \includegraphics[width=\textwidth]{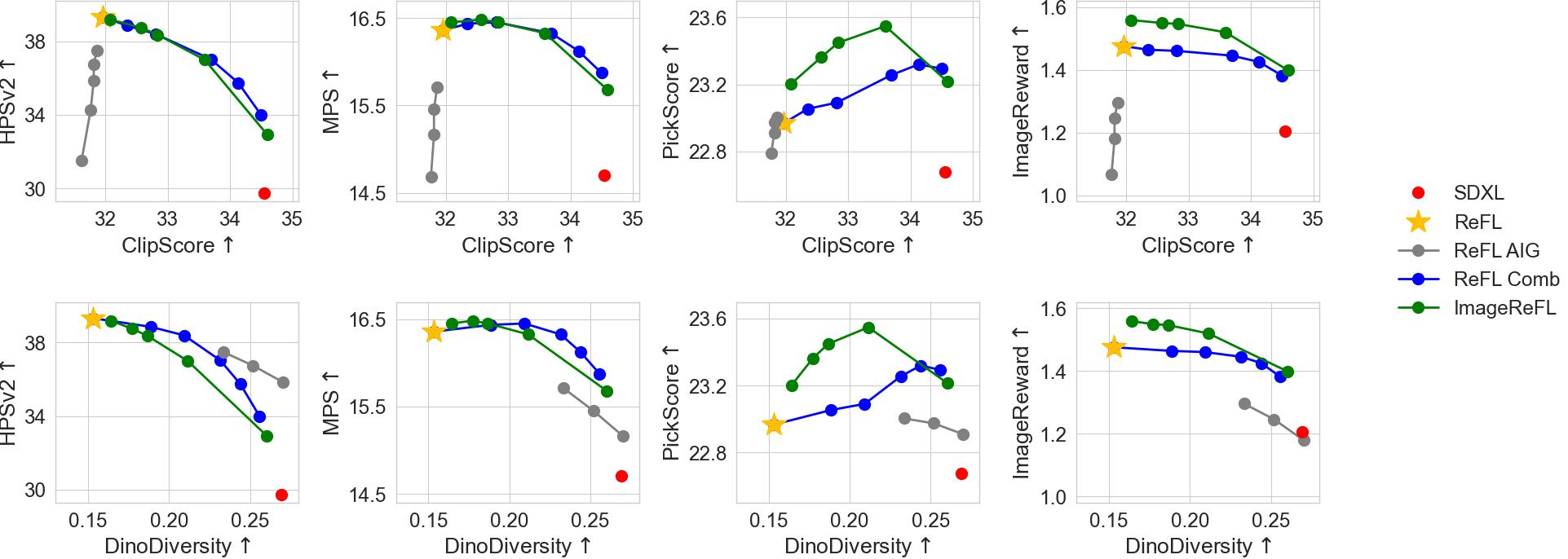}
    \caption{Quality/diversity trade-off for SDXL trained on HPSv2.1. ImageReFL outperforms both ReFL and ReFL AIG baselines in PickScore and ImageReward, while achieving comparable results in HPSv2.1 and MPS.}
    \label{fig:fig_10}
\end{figure}

\subsection{User Preference Study}

We conducted a user study to validate the advantages of ImageReFL compared to ReFL and ReFL with AIG generation in terms of text alignment, diversity, and image quality. The study involved $20$ participants and collected a total of $4{,}800$ votes across all evaluation criteria. The results are aggregated in Table~\ref{tab:user_study}. Additional details, including the ablation format, are provided in Appendix~\ref{sec:ablation_user_study}. ImageReFL outperforms all baselines in diversity and text-image alignment across experimental setups, and achieves higher visual appeal in all configurations except ReFL with SD 1.5.

\section{Conclusion}\label{sec:conclusion}
We address reward hacking in diffusion model fine-tuning. Our analysis, supported by metrics and a user study, shows that ReFL leads to diversity collapse, while AIG preserves diversity at the cost of image quality. To resolve this, we propose a hybrid method that uses the base model in early diffusion steps and introduce ImageReFL, which fine-tunes later steps to enhance quality.
Our methods outperform ReFL and ReFL AIG in both quantitative metrics and human evaluations. Furthermore, the proposed approach is adaptable and may enhance other techniques such as DRTune and DraftK. A primary limitation is the requirement to maintain two sets of model weights during generation, which can be impractical in resource-constrained settings, particularly with large models. Another constraint is the strong reliance on the reward model: inaccuracies in the reward signal may introduce artifacts into the generated images. However, ImageReFL demonstrates greater robustness to this issue compared to ReFL. 


\newpage

\bibliographystyle{plain}
\bibliography{align_paper}


\newpage
\appendix
\section{Appendix}

\label{sec:ablation_user_study}
\subsection{User study settings}
In the user study, annotators were asked to evaluate generated images based on three criteria:

\begin{enumerate}[label=\textbf{\arabic*.}]
    \item \textbf{Which row of images is more diverse?} \par
    \textit{To what extent do the images within a single row differ from each other in terms of content, colors, composition, and other visual elements?}
    
    \item \textbf{Which row of images corresponds better to the textual description?} \par
    \textit{To what degree do the images reflect the content described in the textual prompt above?}
    
    \item \textbf{Which row of images is more visually appealing?} \par
    \textit{Which images are more aesthetically pleasing in terms of visual quality, design, and overall presentation?}
\end{enumerate}

\begin{figure}[h!]
    \includegraphics[width=\textwidth]{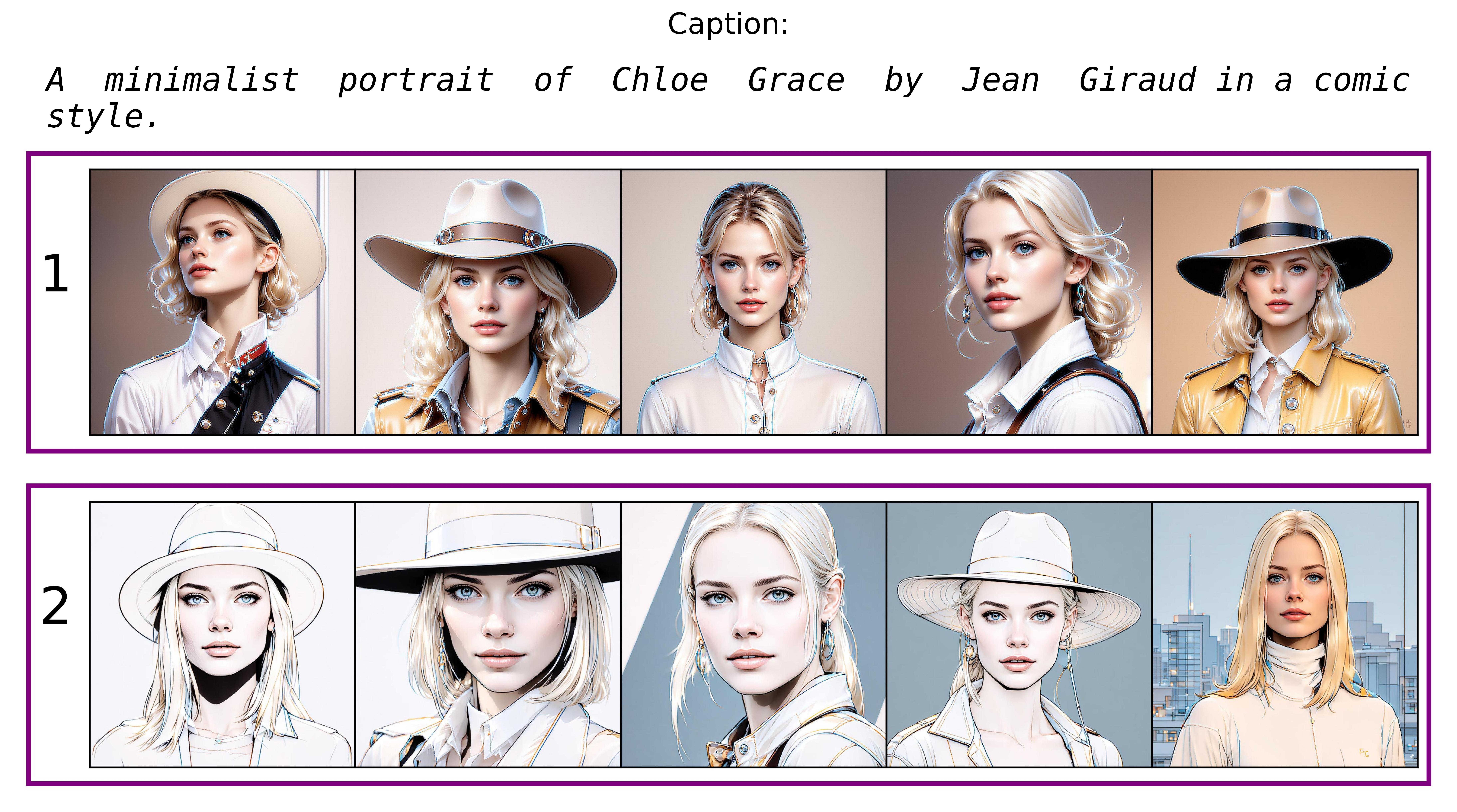}
    \caption{User study image example.}
    \label{fig:fig_us}
\end{figure}

Figure~\ref{fig:fig_us} shows an example task from the study.

For each question, annotators could choose between three options: preference for the first row, preference for the second row, or no preference.

We generated examples for the user study using the test split of the Human Preference Dataset v2, which contains 400 diverse captions. To reduce bias, the rows were randomly swapped. Each task was annotated by three different individuals, and their responses were aggregated for analysis.

\newpage
\label{sec:appendix_pick_score_results}
\subsection{Additional quantitative and qualitative results}
\FloatBarrier

We conducted experiments using two reward models—HPSv2.1 and PickScore—and two base diffusion models: Stable Diffusion 1.5 (SD1.5) and Stable Diffusion XL (SDXL).

Table~\ref{tab:all_metrics_table} presents the evaluation metrics for the best-performing configuration of each method. For SD1.5, we use 10 steps with the base model and 30 steps with the fine-tuned model in the ReFL setup, and 30 steps with the base model and 10 with the fine-tuned model in ImageReFL. For SDXL, the corresponding step configurations are 20/20 for ReFL and 35/5 for ImageReFL. For AIG, we set $\lambda=1$ in all experiments to match ImageReFL's diversity as measured by DinoDiversity, our primary diversity metric.

\begin{figure}[H]
    \includegraphics[width=\textwidth]{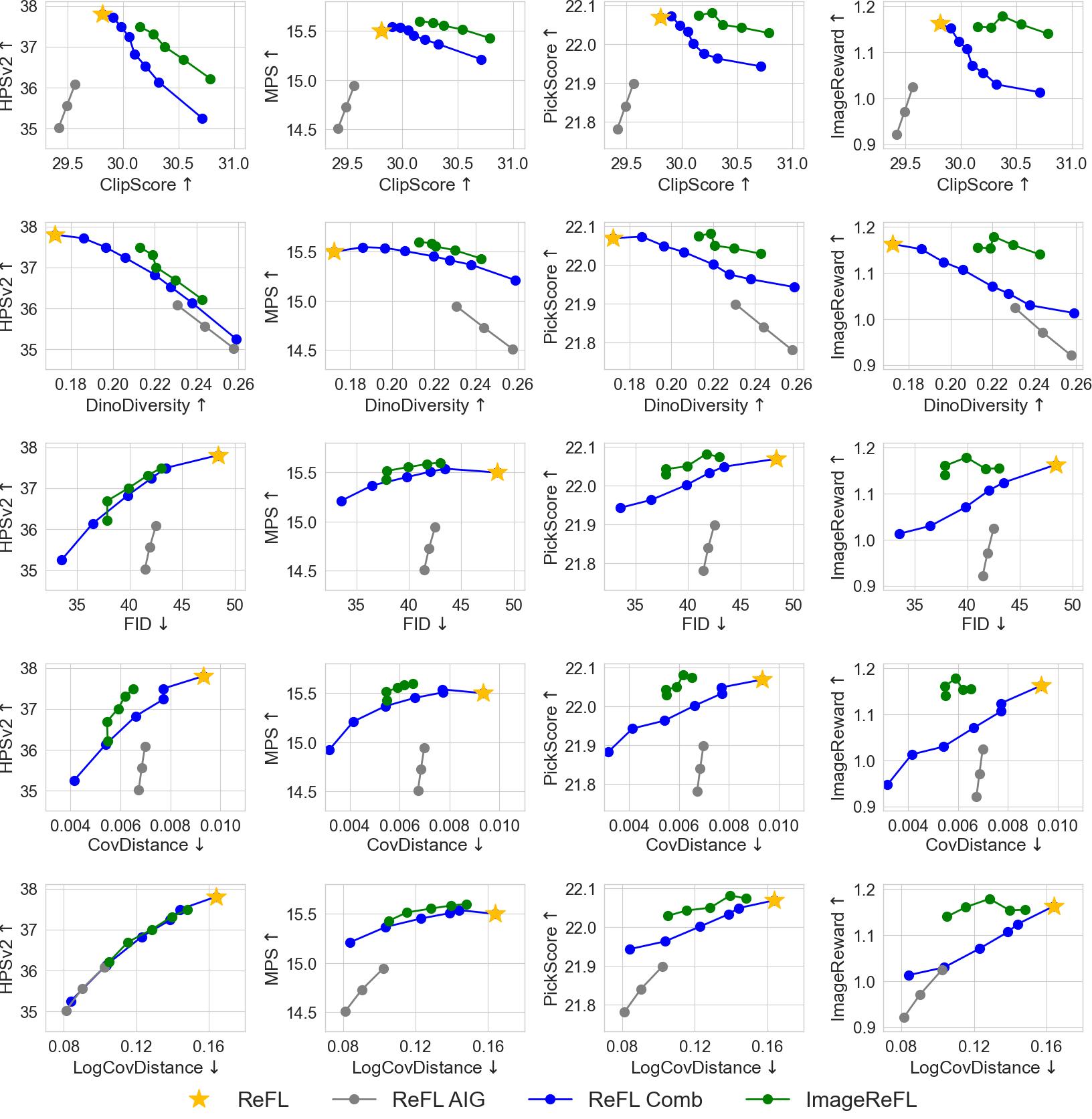}
    \caption{Quality/diversity trade-off for SD1.5 trained on HPSv2.1. ImageReFL outperforms both baselines and achieves a more favorable balance between image quality and diversity.}
    \label{fig:a_fig_1}
\end{figure}

\begin{figure}[H]
    \includegraphics[width=\textwidth]{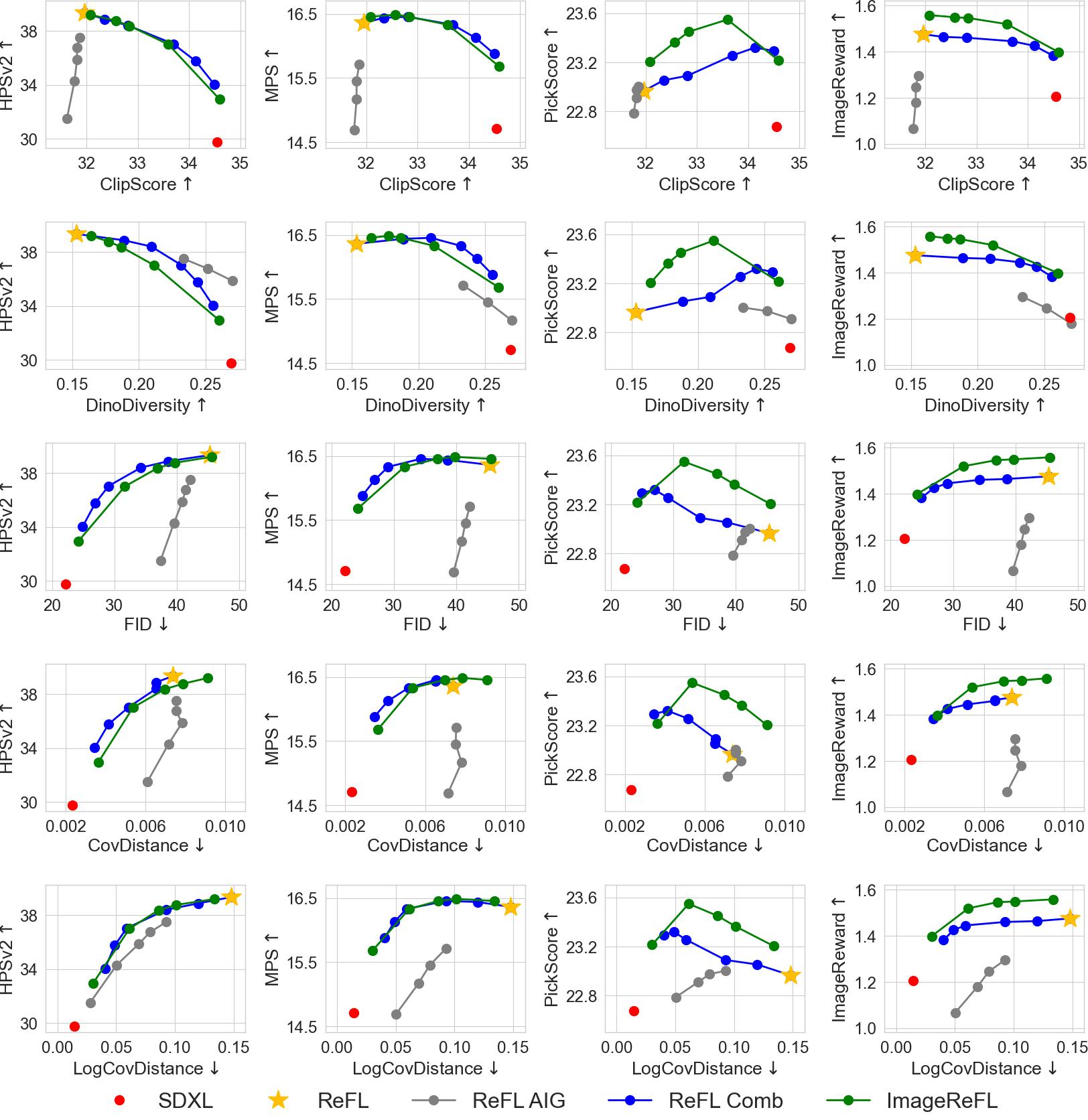}
    \caption{Quality/diversity trade-off for SDXL trained on HPSv2.1. ImageReFL outperforms both ReFL and ReFL AIG baselines in PickScore and ImageReward, while achieving comparable results in HPSv2.1 and MPS.}
    \label{fig:a_fig_2}
\end{figure}

\begin{figure}[H]
    \centering
    \includegraphics[width=\textwidth]{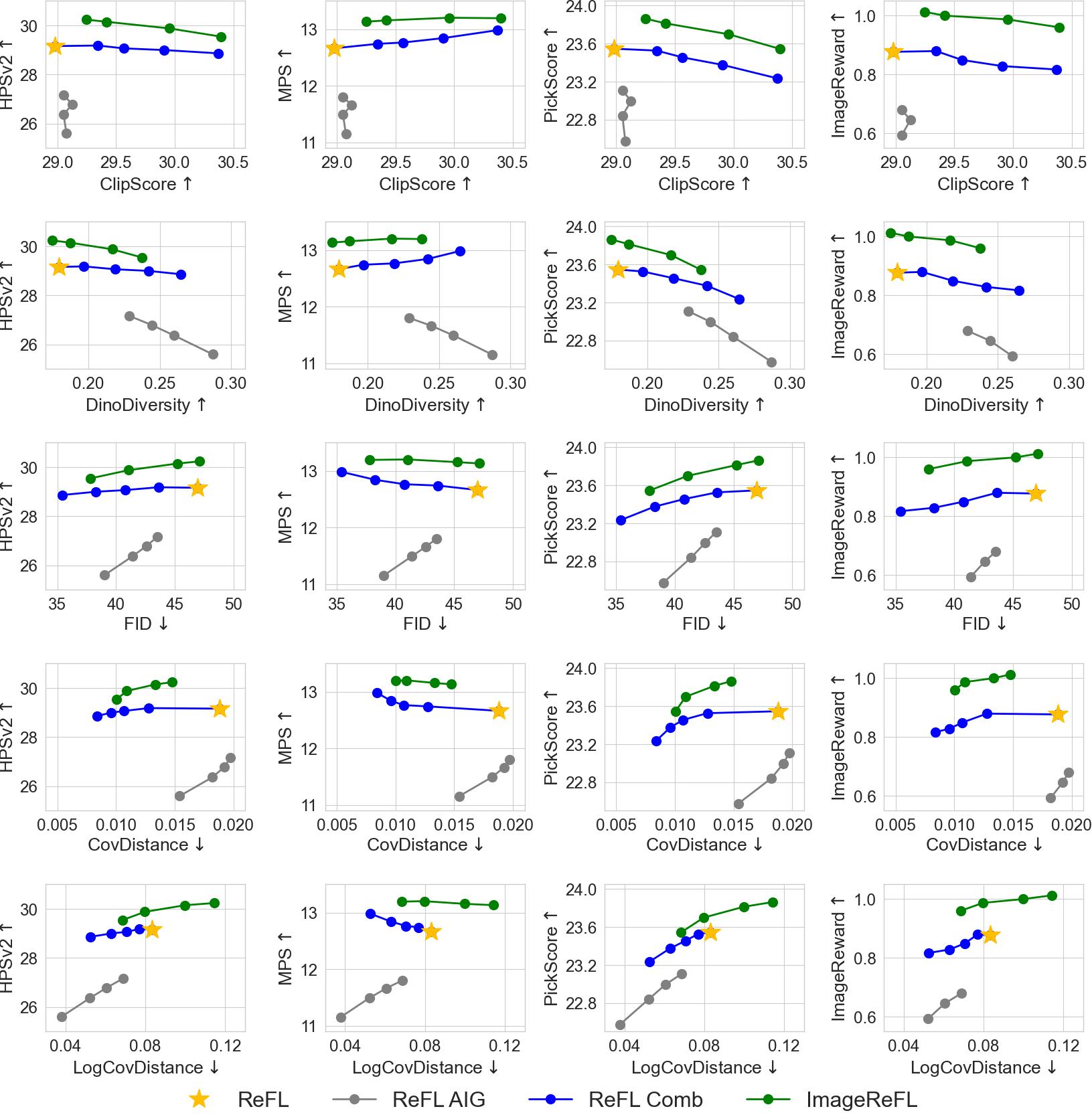}
    \caption{Quality/diversity trade off for SD1.5 trained on PickScore. ImageReFL outperforms
both baselines and achieves a more favorable balance between image quality and diversity.}
    \label{fig:a_fig_3}
\end{figure}

\begin{figure}[H]
    \centering
    \includegraphics[width=\textwidth]{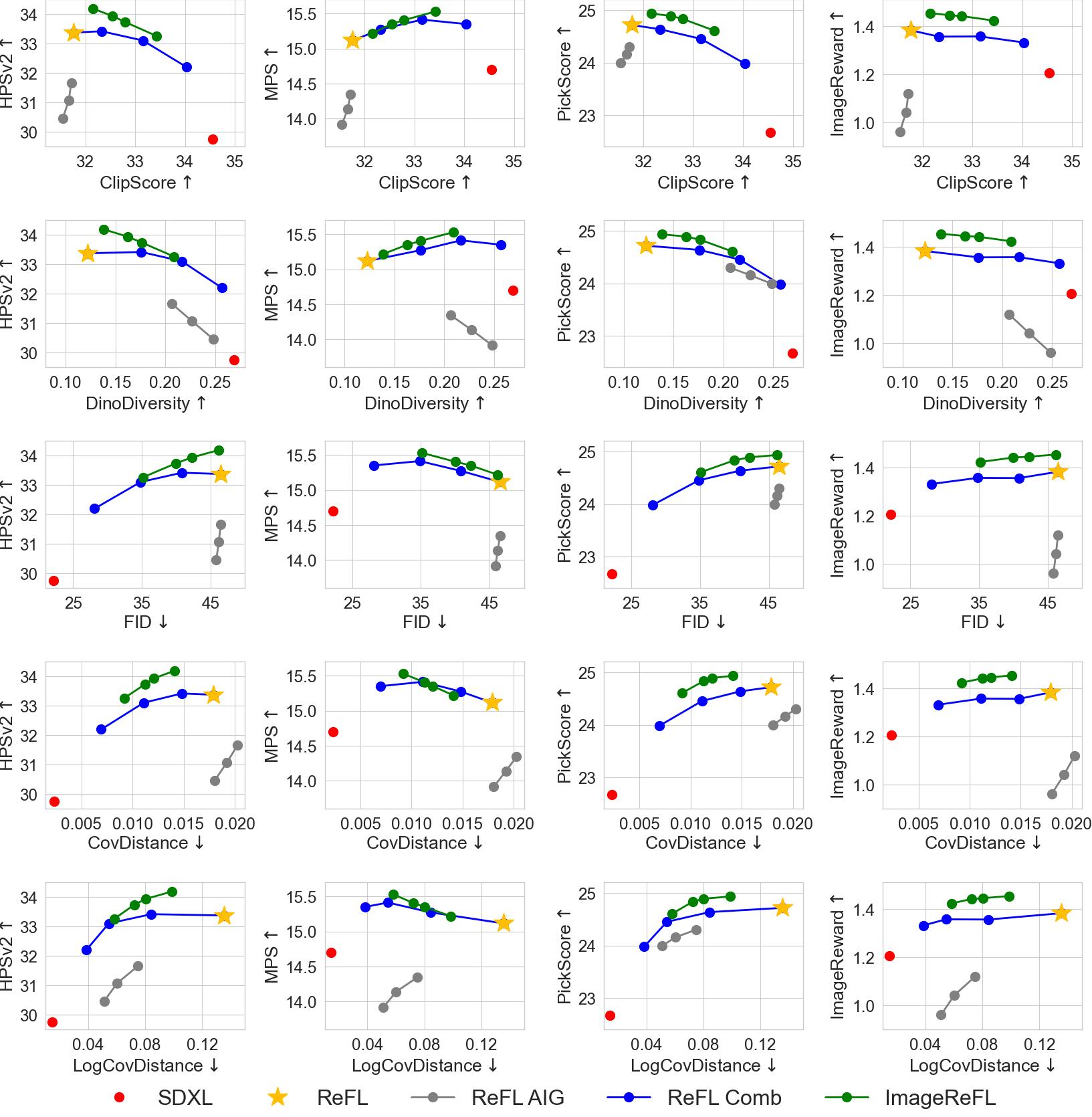}
    \caption{Quality/diversity trade off for SDXL trained on PickScore. ImageReFL outperforms
both baselines and achieves a more favorable balance between image quality and diversity.}
    \label{fig:a_fig_4}
\end{figure}

Figures~\ref{fig:a_fig_1}, \ref{fig:a_fig_2}, \ref{fig:a_fig_3}, and \ref{fig:a_fig_4} illustrate the quality-diversity trade-offs across all experimental setups.

\begin{figure}[H]
    \centering
    \includegraphics[width=0.95\textwidth]{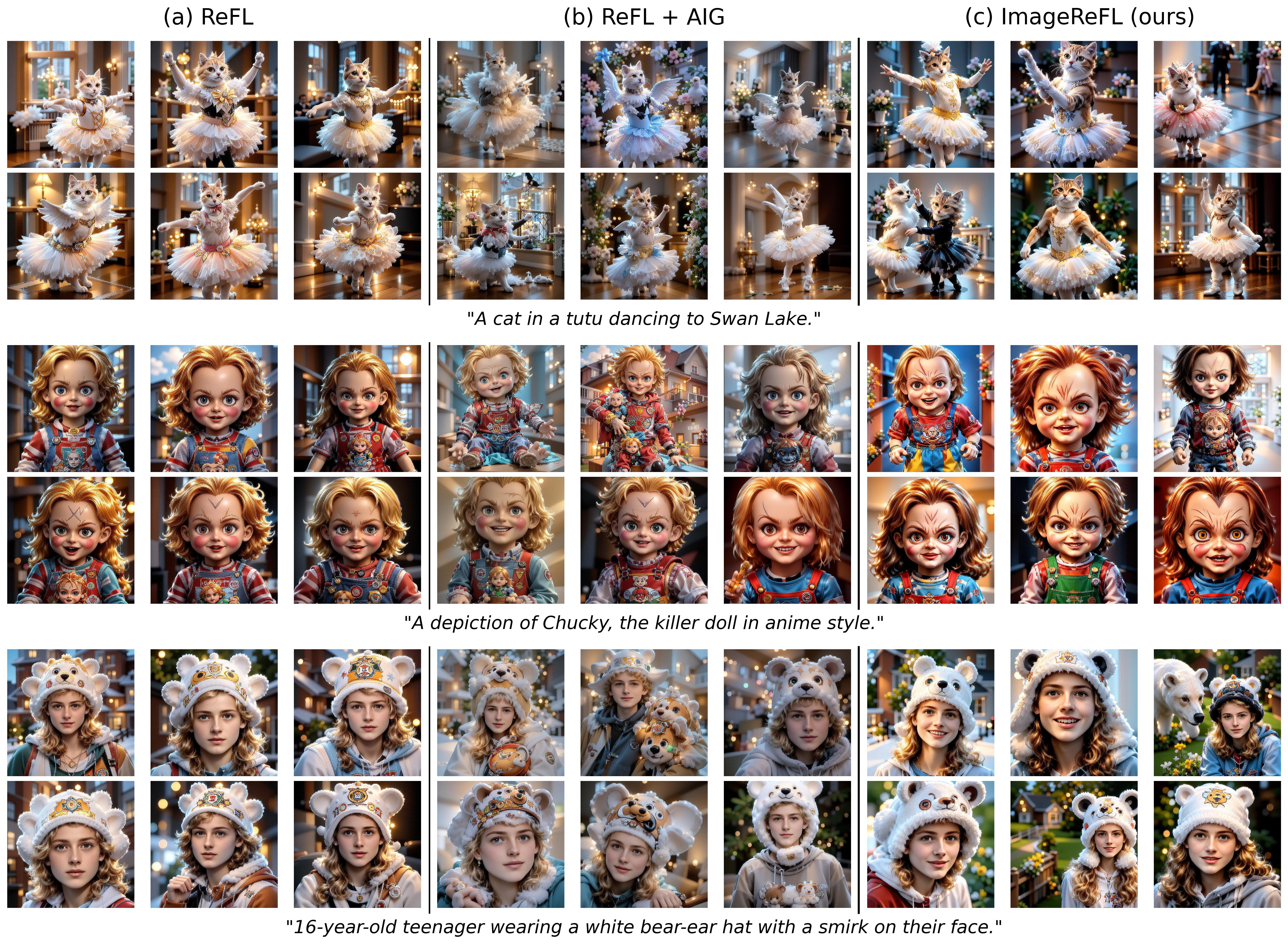}
    \caption{Comparison of ImageReFL with ReFL and ReFL AIG baselines using SD1.5 trained on HPSv2.1.}
    \label{fig:a_fig_7}
\end{figure}

\begin{figure}[H]
    \centering
    \includegraphics[width=0.95\textwidth]{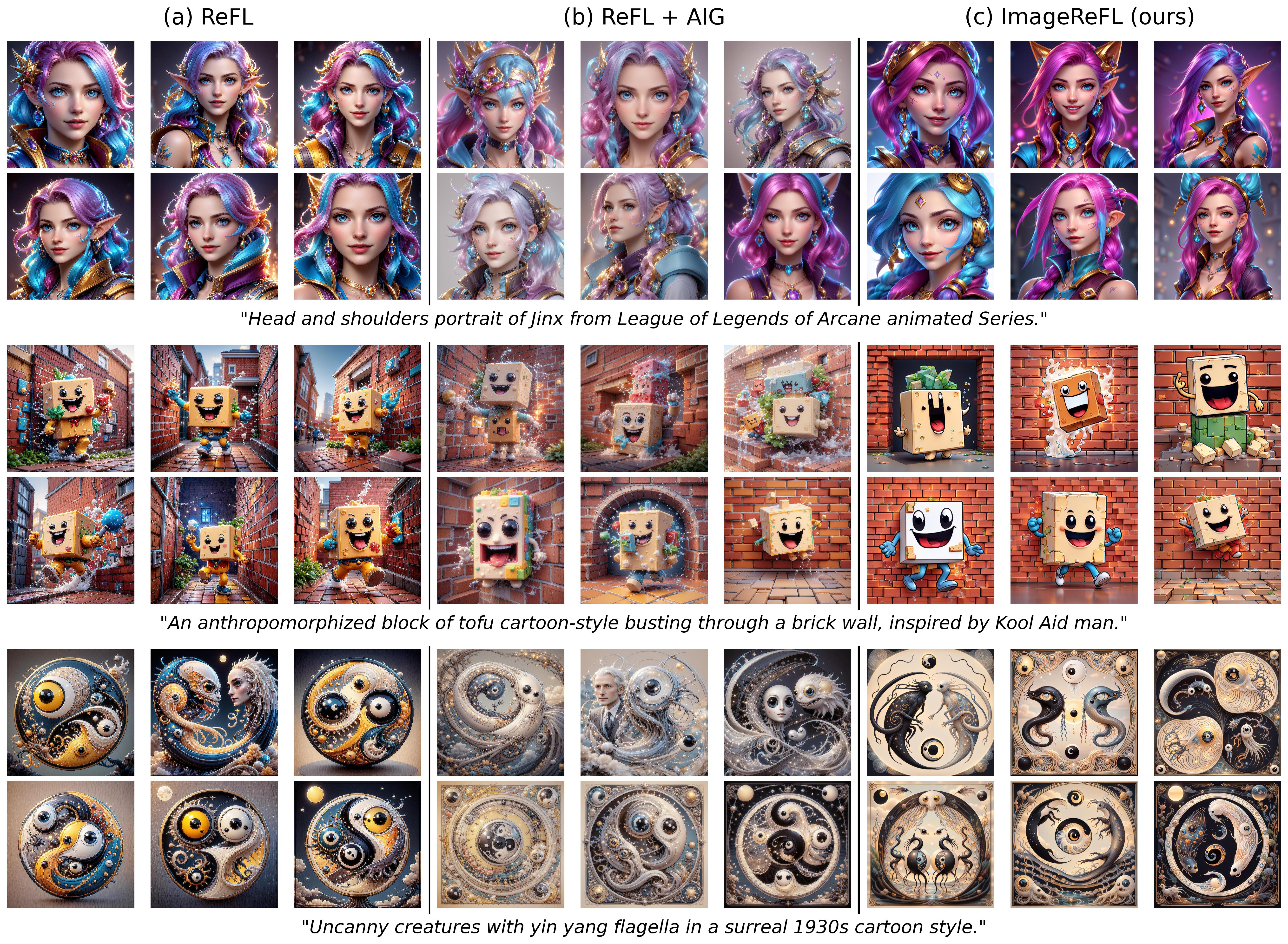}
    \caption{Comparison of ImageReFL with ReFL and ReFL AIG baselines using SDXL trained on HPSv2.1.}
    \label{fig:a_fig_8}
\end{figure}

\begin{figure}[H]
    \centering
    \includegraphics[width=0.95\textwidth]{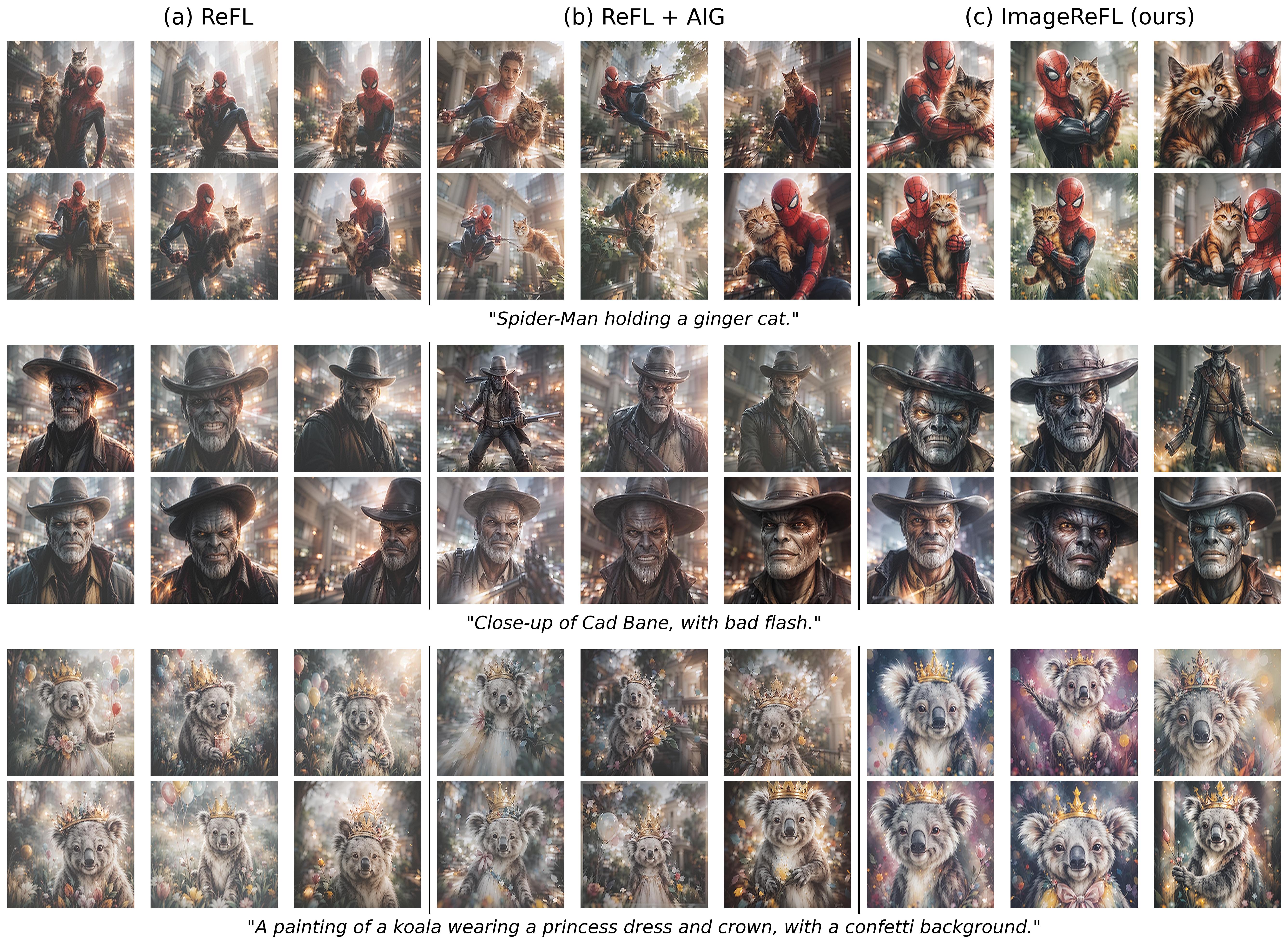}
    \caption{Comparison of ImageReFL with ReFL and ReFL AIG baselines using SD1.5 trained on PickScore.}
    \label{fig:a_fig_11}
\end{figure}

\begin{figure}[H]
    \centering
    \includegraphics[width=0.95\textwidth]{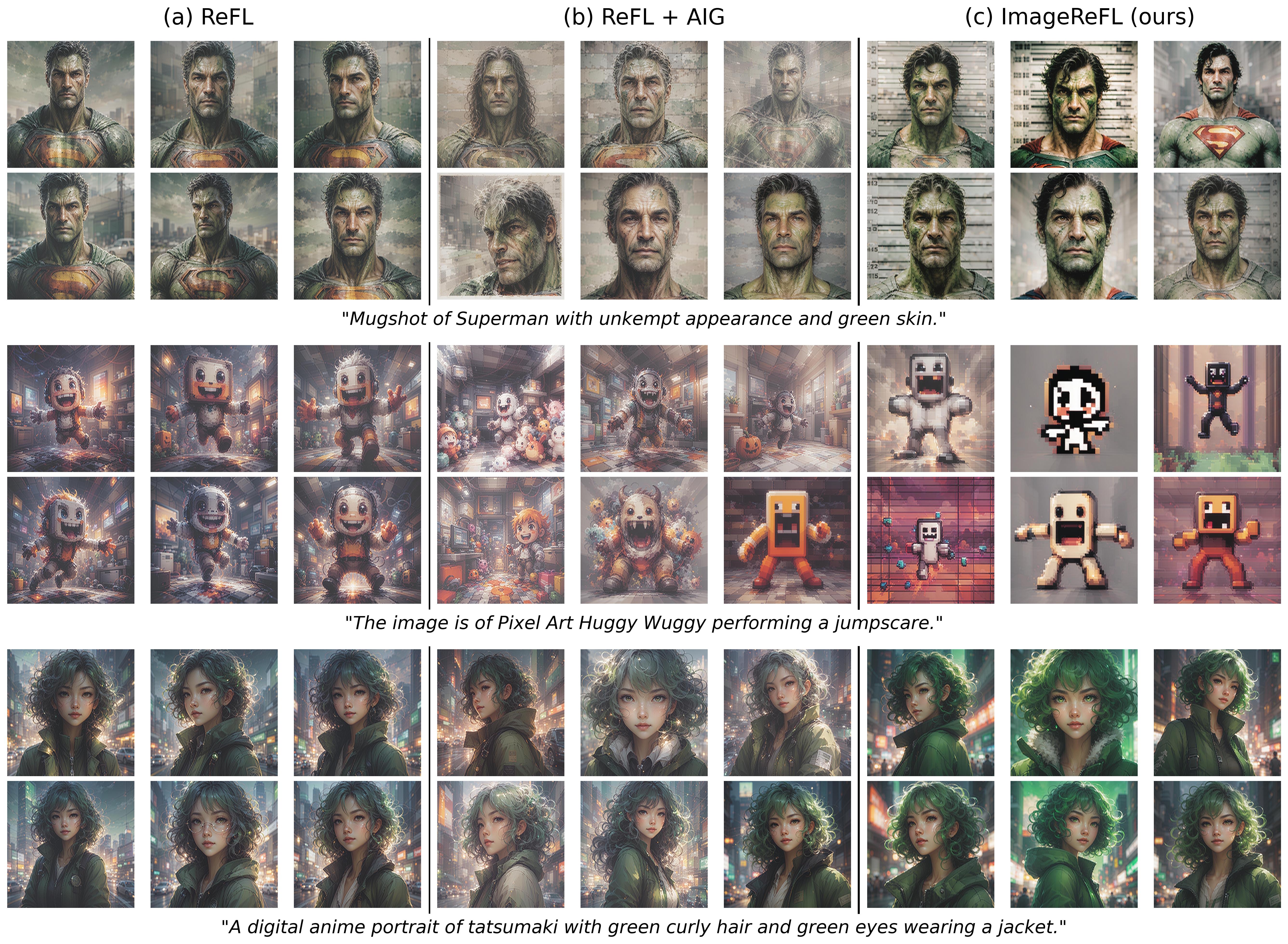}
    \caption{Comparison of ImageReFL with ReFL and ReFL AIG baselines using SDXL trained on PickScore.}
    \label{fig:a_fig_14}
\end{figure}

These extended results support the main findings: both ImageReFL and ReFL with combined generation consistently enhance image diversity without sacrificing quality. Among the methods, ImageReFL often provides the most favorable balance between diversity and visual fidelity. Figures~\ref{fig:a_fig_7} \ref{fig:a_fig_8} \ref{fig:a_fig_11} \ref{fig:a_fig_14} demonstrate examples of image generation.

\begin{table}[h]
    \centering
    \caption{Unified image generation comparison across SD1.5 and SDXL trained with HPSv2.1 and PickScore.}
    \label{tab:all_metrics_table}
    \resizebox{\linewidth}{!}{
    \begin{tabular}{@{}llccccccccc@{}}
    \toprule
    \multicolumn{9}{c}{\textbf{Quality}} & {\textbf{Diversity}}\\
    \cmidrule(lr){3-7} \cmidrule(lr){8-11}
    Setting & Method &
    HPSv2.1 $\uparrow$ &
    MPS $\uparrow$ &
    PickScore $\uparrow$ &
    IR $\uparrow$ &
    ClipScore $\uparrow$ &
    DinoDiv $\uparrow$ &
    FID $\downarrow$ &
    CovDist $\downarrow$ &
    LogCovDist $\downarrow$ \\
    
    & & $\pm$0.07 & $\pm$0.10 & $\pm$0.03 & $\pm$0.018 & $\pm$0.09 
    & $\pm$0.003 & $\pm$0.5 & $\pm$0.0002 & $\pm$0.001 \\
    \midrule

    \shortstack[l]{SD1.5} \\
    HPSv2.1
    & ReFL & \textbf{37.80} & \underline{15.50} & \textbf{22.07} & 1.163 & 29.81 & 0.172 & 48.4 & 0.0093 & 0.164 \\
    & ReFL AIG & 36.07 & 14.94 & 21.89 & 1.024 & 29.56 & \bf{0.231} & 42.5 & 0.0070 & \bf{0.102} \\
    \cmidrule(lr){2-11}
    & ReFL Comb & 36.83 & 15.45 & 22.00 & \underline{1.070} & \underline{30.10} & \underline{0.220} & \textbf{39.83} & \underline{0.0066} & \underline{0.123} \\
    & ImageReFL & \underline{37.00} & \textbf{15.55} & \underline{22.05} & \textbf{1.178} & \textbf{30.37} & \underline{0.220} & \underline{39.91} & \textbf{0.0059} & 0.129 \\

    \midrule
    \shortstack[l]{SD1.5} \\
    PickScore
    & ReFL & \underline{29.16} & 12.66 & \textbf{23.55} & \underline{0.877} & 28.98 & 0.1794 & 47.0 & 0.0188 & 0.083 \\
    & ReFL AIG & 27.16 & 11.80 & 23.10 & 0.680 & 29.05 & \bf{0.229} & 43.5 & 0.020 & \underline{0.069} \\
    \cmidrule(lr){2-11}
    & ReFL Combined & 29.06 & \underline{12.76} & 23.45 & 0.850 & \underline{29.56} & \underline{0.2185} & \underline{40.8} & \underline{0.0107} & 0.071 \\
    & ImageReFL & \textbf{29.54} & \textbf{13.19} & \underline{23.54} & \textbf{0.960} & \textbf{30.39} & 0.2166 & \textbf{37.8} & \textbf{0.0100} & \textbf{0.068} \\

    \midrule
    \shortstack[l]{SDXL} \\
    HPSv2.1
    & ReFL & \textbf{39.31} & \textbf{16.36} & 22.96 & \underline{1.475} & 31.96 & 0.1532 & 45.3 & 0.0073 & 0.147 \\
    & ReFL AIG & \underline{37.48} & 15.71 & 23.00 & 1.295 & 31.86 & \bf{0.233} & 42.16 & 0.0075 & 0.093 \\
    \cmidrule(lr){2-11}
    & ReFL Combined & 37.02 & \underline{16.32} & \underline{23.25} & 1.445 & \textbf{33.69} & \underline{0.2317} & \textbf{29.0} & \textbf{0.0052} & \textbf{0.059} \\
    & ImageReFL & 37.00 & \underline{16.32} & \textbf{23.54} & \textbf{1.519} & \underline{33.59} & 0.2116 & \underline{31.7} & \underline{0.0054} & \underline{0.061} \\

    \midrule
    \shortstack[l]{SDXL} \\
    PickScore
    & ReFL & \textbf{33.38} & 15.11 & \textbf{24.71} & \underline{1.382} & 31.17 & 0.1223 & 47.0 & 0.0179 & 0.135 \\
    & ReFL AIG & 31.66 & 14.34 & 24.30 & 1.120 & 31.71 & 0.206 & 46.5 & 0.0202 & 0.075 \\
    \cmidrule(lr){2-11}
    & ReFL Combined & 33.10 & \underline{15.41} & 24.45 & 1.357 & \underline{33.15} & \textbf{0.2166} & \textbf{34.8} & \underline{0.0111} & \textbf{0.054} \\
    & ImageReFL & \underline{33.26} & \textbf{15.53} & \underline{24.60} & \textbf{1.423} & \textbf{33.34} & \bf{0.2090} & \underline{35.1} & \textbf{0.0092} & \underline{0.058} \\

    \bottomrule
    \end{tabular}
}
\end{table}
\hfill
\FloatBarrier

\end{document}